\documentclass[10pt,twocolumn,letterpaper]{article}
\usepackage{wacv}
\wacvfinalcopy
\usepackage{times}
\usepackage{epsfig}
\usepackage{graphicx}
\usepackage{amsmath}
\usepackage{amssymb}
\usepackage{multirow}
\usepackage{mathtools}
\usepackage{algorithmic}
\usepackage{algorithm}
\usepackage[inline]{enumitem}
\newcommand*{\Resize}[2]{\resizebox{#1}{!}{$#2$}}%
\DeclarePairedDelimiterX{\infdivx}[2]{(}{)}{%
  #1\;\delimsize\|\;#2%
}
\newcommand{\infdiv}{\infdivx}
\DeclarePairedDelimiter{\norm}{\lVert}{\rVert}
\usepackage{subcaption}
\usepackage{comment} 
\usepackage{tabto}

\def\wacvPaperID{264} 

\ifwacvfinal
\usepackage[breaklinks=true,bookmarks=false]{hyperref}
\else
\usepackage[pagebackref=true,breaklinks=true,colorlinks,bookmarks=false]{hyperref}
\fi

\ifwacvfinal
\else
\fi

\begin{document}

\title{Preventing Catastrophic Forgetting and Distribution Mismatch in Knowledge Distillation via Synthetic Data}

\author{Kuluhan Binici$^{\ast \dagger}$, Nam Trung Pham$^{\ast}$, Tulika Mitra$^{\dagger}$, Karianto Leman$^{\ast}$\\
$^{\ast}$Institute for Infocomm Research, A*STAR, Singapore \\
        $^{\dagger}$School of Computing, National University of Singapore, Singapore\\
{\tt\small \{kuluhan, tulika\}@comp.nus.edu.sg, \{ntpham, karianto\}@i2r.a-star.edu.sg}\\

}

\maketitle
\thispagestyle{empty}

\begin{abstract}
With the increasing popularity of deep learning on edge devices, compressing large neural networks to meet the hardware requirements of resource-constrained devices became a significant research direction.
Numerous compression methodologies are currently being used to reduce the memory sizes and energy consumption of neural networks. Knowledge distillation (KD) is among such methodologies and it functions by using data samples to transfer the knowledge captured by a large model (teacher) to a smaller one (student).
However, due to various reasons, the original training data might not be accessible at the compression stage. Therefore, data-free model compression is an ongoing research problem that has been addressed by various works. 
In this paper, we point out that catastrophic forgetting is a problem that can potentially be observed in existing data-free distillation methods. Moreover, the sample generation strategies in some of these methods could result in a mismatch between the synthetic and real data distributions. To prevent such problems, we propose a data-free KD framework that maintains a dynamic collection of generated samples over time. Additionally, we add the constraint of matching the real data distribution in sample generation strategies that target maximum information gain. Our experiments demonstrate that we can improve the accuracy of the student models obtained via KD when compared with state-of-the-art approaches on the SVHN, Fashion MNIST and CIFAR100 datasets.
\end{abstract}

    \begin{figure*}[ht!]
        \centering
        {\includegraphics[height=.28\textwidth]{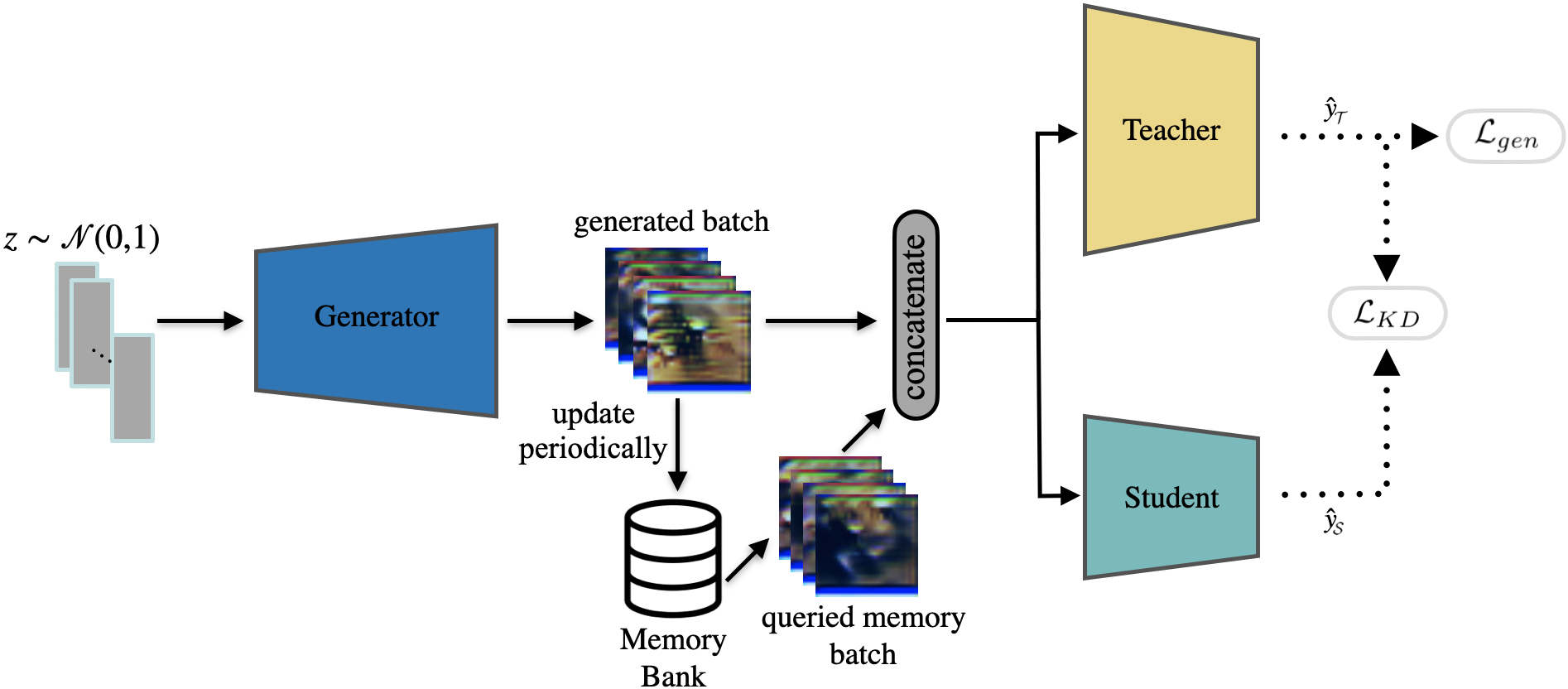}}\hfill
        \caption{Proposed Data-Free Knowledge Distillation framework. At each iteration, the student is frozen and the generator is trained to produce samples that are confidently classified by the teacher but not by the student. Later the generator is frozen and the student is trained by a combination of samples inferred from the generator and retrieved from the memory bank. The bank is updated with the newly generated samples at a pre-determined frequency.}
        \label{fig:method}
    \end{figure*}

\vspace{-12pt}
\section{Introduction}
\label{sec:intro} 
Large-scale deep learning models have achieved overwhelming success over many different learning tasks~\cite{reed2016generative, long2015fully, radford2019language, carreira2017quo}. However, the huge computational complexities and massive storage requirements make it challenging to deploy them on edge devices with low power and storage capacities such as mobile phones. Knowledge Distillation \cite{hinton2015distilling} is a popular method used in compressing complex models and has been proven to be successful in maintaining model performance after compression ~\cite{asami2017domain,shi2019knowledge,sun2019patient}. 
The process can be defined as extracting knowledge from a complex pre-trained neural network, called the teacher, to train a relatively more compact student network.
This extraction is generally done by forcing the student to imitate the responses of the teacher against the training data. Unfortunately, this data dependency can cause problems when the compression 
process is carried out by
groups other than the model developers. In such scenarios, the group that attempts compression might not have access to the training data due to privacy-related issues or they might not have the means to relocate and store the dataset if it is large in volume. Hence, the whole process could become infeasible. 
\par
Recognition of this problematic coupling of KD with data has recently attracted attention from the scientific community ~\cite{micaelli2019zero, choi2020data, luo2020large}. 
A common way to address this problem is to use synthetic samples for distillation. This task is often referred to as \textit{data-free knowledge distillation} in the literature. Most of the existing works utilize a neural network (generator) to generate synthetic samples. The network parameters are updated throughout distillation via feedback from the teacher network and/or teacher-student performance gap (see Section \ref{sec:DF_KD}). Although the reported results are promising, they still need improvement for data-free methods to become the standard way of compressing image models. There are two aspects of the current data synthesis strategies that we identified as performance bottlenecks. 
The first one is that, using only newly generated samples to train the student after each time generator's weights are updated  
~\cite{chen2019data,fang2019data,han2021robustness,luo2020large, micaelli2019zero}, could cause the student network to forget the knowledge it acquired in the earlier steps. The reason is that, unlike in regular KD, the synthetic data distribution changes over time as the generator is updated in data-free methods. Therefore, performing KD only with a freshly generated batch, without having stored samples from earlier iterations, could cast the student vulnerable against such distribution shifts.
Secondly, targeting the generation of samples over which the student and the teacher have maximum disagreement ~\cite{fang2019data, micaelli2019zero}, could yield a student that is optimal for a different distribution than the original data. 
\par
To address the above-mentioned performance bottlenecks, we propose a memory mechanism to store generated samples over iterations and a generator objective that targets samples that both maximize teacher-student disagreement and approximate real data distribution.
Our contributions in this work have three folds and can be summarized as:
\begin{itemize}
    \item Identification of the catastrophic forgetting problem as a performance bottleneck for available KD methods using synthetic data.
    \item A data-free KD framework that mitigates catastrophic forgetting by keeping memory of generated samples over iterations. 
    \item Preventing possible mismatches between the generated and the original data distributions with an enhanced sample generation strategy that improves upon the state-of-the-art. 
\end{itemize}
\section{Related Works}
\label{sec:rel_work}
\subsection{Catastrophic Forgetting}
\label{sec:cat_forgetting}
The term \textit{Catastrophic Forgetting} was defined by French \cite{french1999catastrophic} to describe the loss of previously learned information observed in neural networks when they are sequentially trained to learn new information. Unlike natural cognitive systems, neural networks cannot retain the knowledge they had previously acquired while they keep learning from new data. To preserve such knowledge, they should also be exposed to old data at every update. Goodfellow et al. \cite{goodfellow2013empirical} investigated the extent to which catastrophic forgetting affects neural networks. To alleviate these effects, Kirkpatrick et al. \cite{kirkpatrick2017overcoming} proposed to selectively slow down the learning rates of the weights associated with the previously learned information.
\subsection{Data-Free Knowledge Distillation}
\label{sec:DF_KD}
There are two main synthetic data generation approaches in existing data-free KD works. 
\par
The first one is called {\em model inversion} and can be described as inverting the information flow in a neural network to reconstruct appropriate input samples based on the imposed constraints. Nayak et al. \cite{nayak2019zero} proposed to account for the correlations between target classes while generating samples. They used these correlations to construct soft target labels that describe how synthetic samples should be. Yin et al. \cite{yin2020dreaming} used one-hot target labels and applied the Jensen-Shannon divergence between the model-to-be-inverted (teacher) and the student to diversify the synthetic samples. Haroush et al. \cite{haroush2020knowledge} leveraged the batch normalization statistics to make the synthetic data better mimic the real training data. One weakness of the model inversion-based methods is that they require a vast amount of time to create sufficiently large datasets for distillation \cite{yin2020dreaming}. Another one is that some methods include loss terms that constrain the synthetic images to be realistic in the learning objective. The synthesis of realistic images could violate the goal of preserving privacy related to the training data. 
\par
The second approach is to use a generator that, once trained, can produce synthetic data suitable for KD. Yoo et al. \cite{yoo2019knowledge} published an early work in this category and proposed a Variational Auto Encoder (VAE) \cite{kingma2013auto} to generate images that can be recognized by the teacher model. Later, these images are used to distill the knowledge from the teacher to the student. However, by separating the image generation process from KD, this work does not directly attempt to optimize distillation performance. Chen et al.
\cite{chen2019data} used a data-free GAN \cite{goodfellow2014generative} setup and included the distance between the teacher and student responses as a term to be minimized in the objective function. Micaelli and Storkey. \cite{micaelli2019zero}, and Fang et al. \cite{fang2019data} targeted synthetic samples that maximize the information gain for the student. Throughout the paper, we will refer to these samples as "novel" samples. The disagreement between the teacher and the student is used as a proxy to quantify the information gain.

\begin{figure*}[!ht]
\begin{minipage}[b]{.32\linewidth}
  \centering
  \centerline{\includegraphics[height=.48\textwidth]{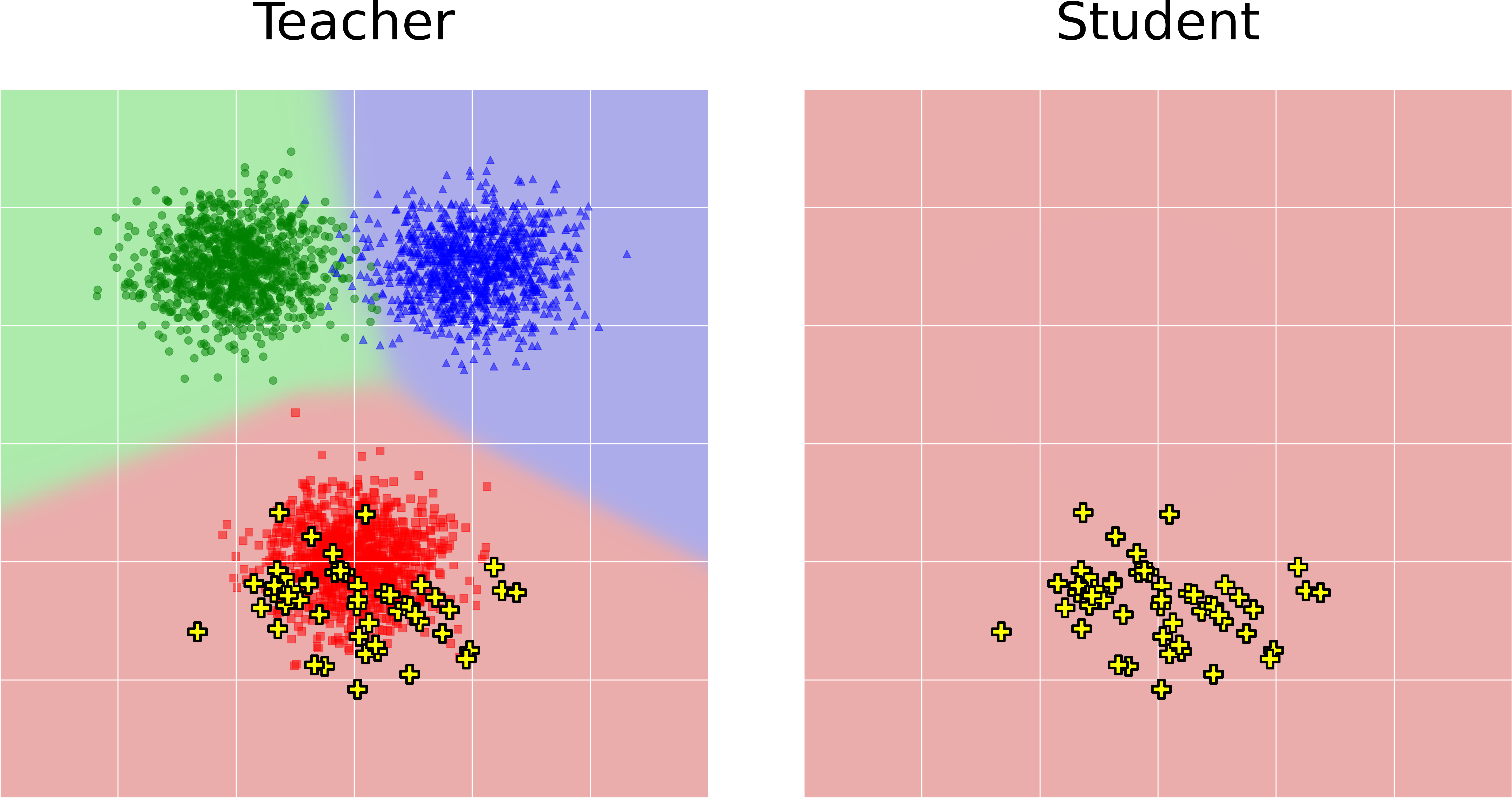}}
  \centerline{Epoch 1}\smallskip
\end{minipage}
\begin{minipage}[b]{.34\linewidth}
  \centering
  \centerline{\includegraphics[height=.48\textwidth]{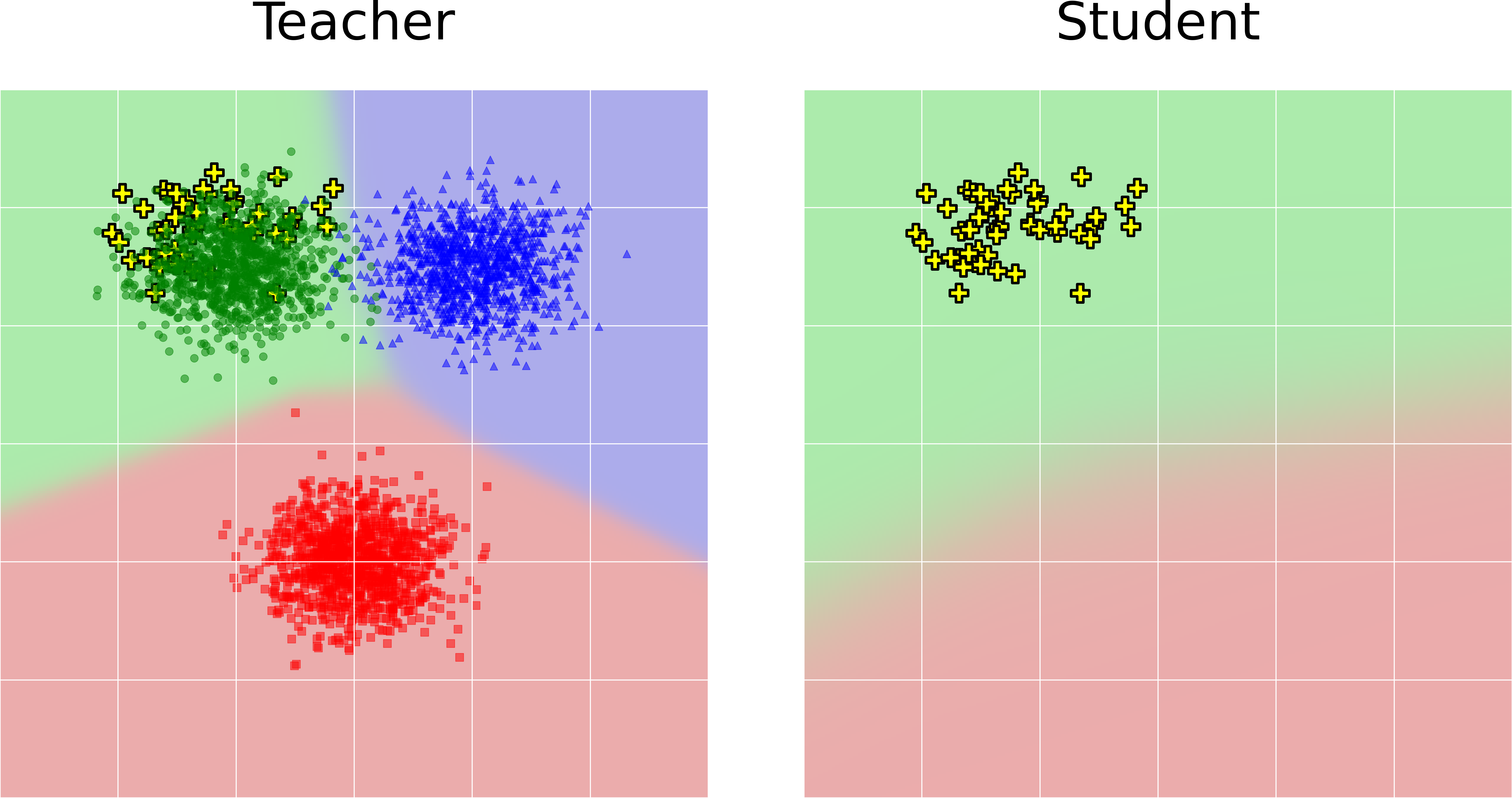}}
   \centerline{Epoch 2}\smallskip
\end{minipage}
\hfill
\begin{minipage}[b]{0.32\linewidth}
  \centering
  \centerline{\includegraphics[height=.48\textwidth]{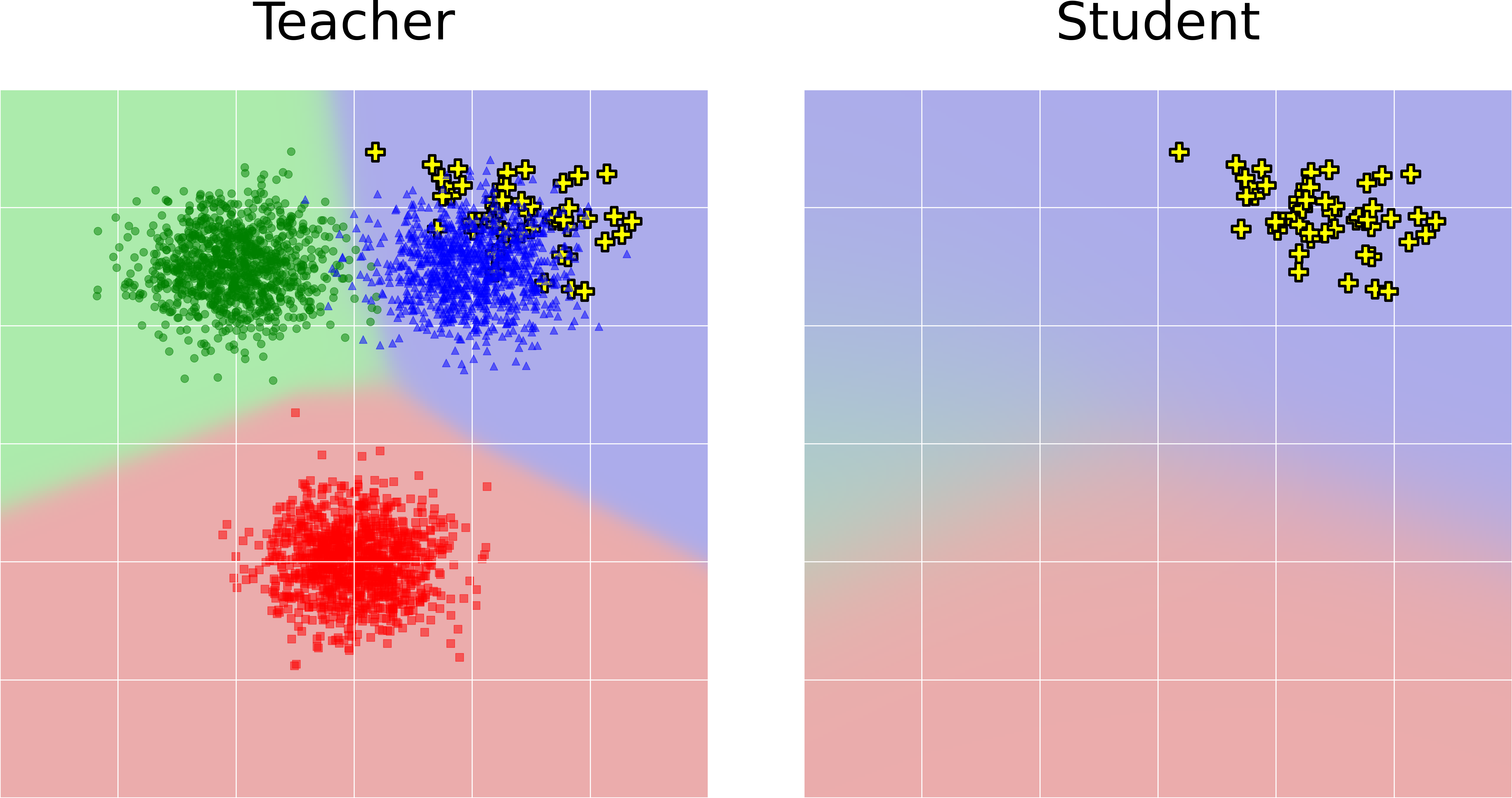}}
  \centerline{Epoch 3}\smallskip
\end{minipage}
\caption{Example of catastrophic forgetting during data-free distillation. The teacher and student decision boundaries are plotted for 3 epochs. Real data samples are displayed in the teacher's decision space and the synthetic samples are marked with yellow crosses.}
\label{fig:cat_forgetting}
\end{figure*}
\section{Method Description}
\label{sec:method}
Our framework contains a generator that produces synthetic samples to distill knowledge from the teacher to the student. We train the generator and the student in two separate stages, alternatively. During the training stage of the generator, the target is to produce samples that are both novel to the student and can be classified by the teacher with high confidence. Later, when the turn comes to the student, it is trained to imitate the responses of the teacher against the generated data. We also include a fixed-sized memory mechanism to remind the student of the information presented to it over time. In Section \ref{sec:met_cat_forget}, we discuss how the generator loss used in \cite{fang2019data} could cause catastrophic forgetting. Later in Section \ref{subsec:mit_cat_forget} we elaborate on our memory mechanism to mitigate catastrophic forgetting. 
Moreover, we use a loss function that conditions the generator to produce 
novel samples that also approximate the real data distribution. In Section \ref{sec:gen_loss}, we construct our proposed loss function by combining the ideas introduced in the papers \cite{fang2019data} and \cite{chen2019data}. The overview of our framework is given in Figure \ref{fig:method} and procedural details are specified in Algorithm \ref{ZAKD}.
\vspace{10pt}
\subsection{Catastrophic Forgetting in Data-Free KD Methods Targeting Novel Synthetic Samples}
\label{sec:met_cat_forget}
In \cite{fang2019data} the optimal generator produces samples that cause the maximum teacher-student disagreement. Formally, the optimal samples at epoch $t$ ($\hat{x}^*_{(t)}$), could be obtained by the Equation \ref{eq6}.
\begin{equation}
\hat{x}^*_{(t)} = \displaystyle \mathop{argmax}_{\hat{x}} D(\mathcal{T}(\hat{x}),\mathcal{S}_{(t)}(\hat{x}))
\label{eq6}
\end{equation}
where $D$ is a function that measures the distance between the outputs of the teacher ($\mathcal{T}$) and the student at epoch $t$ ($\mathcal{S}_{(t)}$). Assuming that $\mathcal{S}_{(t)}$ is optimized for $\hat{x}^*_{(t)}$ after epoch $t$ based on,
\begin{equation}
\mathcal{S}^*_{(t+1)} = \displaystyle \mathop{argmin}_{\mathcal{S}} D(\mathcal{T}(\hat{x}^*_{(t)}),\mathcal{S}_{(t)}(\hat{x}^*_{(t)}))
\label{eq7}
\end{equation}
the $\hat{x}^*_{(t)}$ will yield minimum teacher-student discrepancy in the next epoch $t+1$. Therefore,
\begin{equation}
 \hat{x}^*_{(t)} \cap \hat{x}^*_{(t+1)} = \emptyset
\label{eq8}
\end{equation}
\begin{equation}
 \hat{x}^*_{(t+1)} = \displaystyle \mathop{argmax}_{\hat{x}} D(\mathcal{T}(\hat{x}),\mathcal{S}^*_{(t+1)}(\hat{x}))
\label{eq9}
\end{equation}
Here we emphasize that, once the student is optimized to imitate the teacher for a set of novel samples, the teacher-student disagreement over those samples will diminish.
Therefore, in the next turn, the generator will attempt to produce different samples for which the student still can not imitate the teacher's responses. This can result in catastrophic forgetting as the distribution of our synthetic samples change over iterations causing the student to forget the information gained in earlier time steps. An example case is given in Figure \ref{fig:cat_forgetting}. In the example, the teacher is trained to classify 2-dimensional data points into 3 classes, denoted by red, green, and blue. Later, a student model is trained by the synthetic samples, marked with yellow crosses, at each epoch. The synthetic samples change over epochs and there is no common sample between synthetic sets across epochs. It can be observed that when we train the student with the synthetic samples generated at the 3rd epoch, the knowledge learned earlier about the green class is forgotten. This example visualizes catastrophic forgetting caused by the above-mentioned data synthesis strategy, in an observable space. 
\subsection{Mitigating Catastrophic Forgetting}
\label{subsec:mit_cat_forget}
To prevent such problem, we propose to store some of the samples generated throughout iterations, in a memory bank. Since the number of accumulated memory samples could increase linearly with the arbitrarily selected number of training epochs, we use a fix-sized list to maintain our bank. The bank is updated periodically at a desired rate by inserting generated samples. If the sample list has reached maximum size, we replace a randomly selected batch with the batch of new samples. 
The procedure is described in Algorithm \ref{alg:data_update}. 
\begin{algorithm}[!h]
\caption{Memory Bank Update}
\label{alg:data_update}
\begin{algorithmic}
\STATE \text{\textbf{INPUT:}} A list structure SL that keeps track of recorded samples, batch size B.
\IF {SL is full}
    \STATE $out\_samples \gets SL.remove\_random(B)$ \
    \STATE $REMOVE(out\_samples)$ \
\ENDIF
\STATE sample B vectors (z) from $\mathcal{N}(0,1)$
\STATE $in\_samples \gets \mathcal{G}(z)$\
\STATE $SL.append(in\_samples)$\
\STATE $SAVE(in\_samples)$
\end{algorithmic}
\end{algorithm}
\par
During distillation, if the stored sample list is not empty, a batch of memory samples is selected randomly and combined with freshly generated samples to train the student. 
 
\subsection{Generator Loss to Prevent Mismatch Between Synthetic and Original Data Distributions}
\label{sec:gen_loss}
As mentioned in Section \ref{sec:DF_KD}, targeting the generation of novel samples has demonstrated promising results on several benchmarks ~\cite{fang2019data,micaelli2019zero}. However, this approach fails to maintain high-quality distillation across different benchmarks. We observed this issue after conducting experiments with various new datasets and student architectures. Throughout our experiments, the training/testing performances of the student models were unstable and highly impacted by the hyper-parameter choices. We attribute this to the lack of constraint to generate samples that come from a similar distribution as the training data. Without such constraint, the distillation can result in a student network that is optimized to imitate the teacher for a certain sample subspace which is different from the original training data. Thus, the student might fail to behave like the teacher when presented with samples from the original data distribution. Even if during the distillation, the novel samples coincidentally correspond with real samples, once the student is optimized for them, the next batches of novel samples will be different. If the distillation continues after such correspondence, the student will forget the decision boundaries related to the real data distribution, causing its accuracy on the real test set to degrade. 

To prevent this issue, we propose to generate samples that both induce high discrepancy between teacher and student predictions, and stimulate similar responses from the teacher as the real data.  
Our optimization objective while training the generator, has two main components and can be denoted as, 
\begin{equation}
\mathcal{L}_{gen} = \mathcal{L}_{match} + \mathcal{L}_{d} \label{eq10}
\end{equation}
We give the details of these components in the following sections.
\paragraph{Matching Training Data Distribution}
To generate samples that match the distribution of the original training data, we adopt the optimization objective described in \cite{chen2019data} as 
\begin{equation}
\mathcal{L}_{match} = \mathcal{L}_{oh} + \alpha\mathcal{L}_a + \beta\mathcal{L}_{ie},
\label{eq11}
\end{equation}
where $\alpha$ and $\beta$ are coefficients to adjust the weighted contributions of the participating terms.
$\mathcal{L}_{oh}$ is the one-hot loss that causes the teacher outputs to be one-hot vectors when minimized. It can be defined as the the cross-entropy between the teacher's softmax outputs ($\hat y_{\mathcal{T}}^i=\mathcal{T}(\hat{x}^i))$) and the predicted one-hot labels ($t_{\mathcal{T}}^{i}=argmax(\hat y_{\mathcal{T}}^i)$) when image $\hat{x}^i$ is the input.
\begin{equation}
\mathcal{L}_{oh} = \frac{1}{n}\sum_i t_{\mathcal{T}}^ilog(\hat y_{\mathcal{T}}^i)
\end{equation}
$\mathcal{L}_{a}$ stands for the activation loss and is motivated by the observation of meaningful inputs causing higher valued activation maps in a trained network. The term is defined as
\begin{equation}
\mathcal{L}_{a} = -\frac{1}{n}\sum_i \norm{f_{\mathcal{T}}^i}_1
\end{equation}
where $f_{\mathcal{T}}^i$ denotes the activation values observed at all of the selected layers $i$ of the teacher ($\mathcal{T}$).
\par
Lastly, $\mathcal{L}_{ie}$ is the term that imposes the generator to produce an equal amount of images from each category. When minimized, the entropy of the number of images generated per category gets maximized causing each category to have a similar amount of samples. If we define the probability distribution of class
predictions as $p(\hat y_{\mathcal{T}})=\frac{1}{n}\sum_i \hat y_{\mathcal{T}}^i$, where $y_{\mathcal{T}}^i$ is the softmax output vector for the sample $i$, then the loss term can be denoted as
\begin{equation}
\mathcal{L}_{ie} = -\sum_j p(\hat y_{\mathcal{T}})^jlog(p(\hat y_{\mathcal{T}})^j)
\end{equation}
Here $p(\hat y_{\mathcal{T}})^j$ is the averaged occurrence frequency of the class indexed by $j$ among the generated samples.

\paragraph{Maximum Teacher-Student Discrepancy}
\label{subsec:teacher-student-disc}
Commonly, L1 distance or Kullback-Leibler (KL) divergence between the teacher and student responses are used to quantify disagreement in the loss function. Instead we choose Jensen-Shannon (JS) divergence (Equation \ref{eq5_3}), inspired by \cite{yin2020dreaming}. 
The loss term can be denoted by
\begin{equation}
M = \frac{1}{2}(\mathcal{T}(\hat{x}) + \mathcal{S}(\hat{x}))
\label{eq5_1}
\end{equation}
\begin{equation}
\Resize{7.5cm}{JS(\mathcal{T}(\hat{x}),\mathcal{S}(\hat{x})) = \frac{1}{2}(KL\infdiv{\mathcal{T}(\hat{x})}{M} + \\ KL\infdiv{\mathcal{S}(\hat{x})}{M})}
\label{eq5_2}
\end{equation}
\begin{equation}
\mathcal{L}_{d} = 1 - JS(\mathcal{T}(\hat{x}),\mathcal{S}(\hat{x}))
\label{eq5_3}
\end{equation}
where KL stands for the KL divergence. Our motivation to choose JS divergence to represent the discrepancy between the predictions of the teacher and student, is discussed in section \ref{subsec:ablation}.
\subsection{Knowledge Distillation}
\label{ssec:KD}
In the knowledge distillation phase, the main objective is to get the student responses as close to those of the teacher as possible. To do so, we employ the
L1 distance 
between the responses of the teacher and student networks (Equation \ref{eq8}). An analysis on the impacts of other possible distance terms was given in \cite{fang2019data}. 
\begin{equation} 
\mathcal{L}_{KD} =\norm{t_{\mathcal{T}} - t_{\mathcal{S}}}_1
\label{eq8}
\end{equation}
\begin{algorithm}[!h]
\caption{Data-Free Knowledge Distillation}
\label{ZAKD}
\begin{algorithmic}
\STATE \text{\textbf{INPUT:} A trained teacher network $\mathcal{T}(x; \theta_{\mathcal{T}})$, batch size B}
\STATE Initialize the more compact student $\mathcal{S}(x; \theta_{\mathcal{S}})$ network and the generator $\mathcal{G}(z; \theta_{\mathcal{G}})$
\FOR{number of epochs}
     \FOR{k steps}
        \STATE sample B vectors ($z$) from $\mathcal{N}(0,1)$\
        \STATE $\hat{x} \gets \mathcal{G}(z)$\
        \IF {memory bank is not empty}
            \STATE sample batch $\hat{x}_M$ from memory \
            \STATE $\hat{x} \gets CONCAT(\hat{x}, \hat{x}_M)$\
        \ENDIF
        \STATE $\hat{y}_{\mathcal{T}} \gets \mathcal{T}(\hat{x})$\
        \STATE $\hat{y}_{\mathcal{S}} \gets \mathcal{S}(\hat{x})$\
        \STATE Calculate the loss term $\mathcal{L}_{KD}$ given $\hat{y}_{\mathcal{T}}$ and $\hat{y}_{\mathcal{S}}$
        \STATE $\theta_{\mathcal{S}} \gets optimizer.step(backward(\mathcal{L}_{KD}, \theta_{\mathcal{S}}),\theta_{\mathcal{S}})$
    \ENDFOR
    \STATE sample B vectors ($z$) from $\mathcal{N}(0,1)$\
    \STATE $\hat{x} \gets \mathcal{G}(z)$\
    \STATE $\hat{y}_{\mathcal{T}}, f_{\mathcal{T}} \gets \mathcal{T}(\hat{x})$\
    \STATE $\hat{y}_{\mathcal{S}} \gets \mathcal{S}(\hat{x})$\
    \STATE Calculate the loss terms $\mathcal{L}_{oh}$, $\mathcal{L}_{a}$ , $\mathcal{L}_{ie}$ and $\mathcal{L}_{JS}$ given $\hat{x}$, $\hat{y}_{\mathcal{T}}$, $\hat{y}_{\mathcal{S}}$, $ f_{\mathcal{T}}$
    \STATE $\mathcal{L}_{gen} \gets \mathcal{L}_{oh} + \alpha\mathcal{L}_a + \beta\mathcal{L}_{ie} + \mathcal{L}_{JS}$
    \STATE $\theta_{\mathcal{G}} \gets optimizer.step(backward(\mathcal{L}_{gen}, \theta_{\mathcal{G}}),\theta_{\mathcal{G}})$
    \IF {epoch \% update\_frequency == 0}
        \STATE update memory bank
    \ENDIF
\ENDFOR
\end{algorithmic}
\end{algorithm}
\vspace{-20pt}
\begin{table*}[!ht]
\centering
\begin{tabular}{l|l|l|l|l|l|l|ll}
\hline
                & \multicolumn{2}{c|}{MNIST}         & \multicolumn{2}{c|}{SVHN}          & \multicolumn{2}{c|}{CIFAR10}       & \multicolumn{2}{c}{Fashion MNIST}                       \\ 
                 & \multicolumn{2}{c|}{$\mathcal{T}:$LeNet5}         & \multicolumn{2}{c|}{$\mathcal{T}:$ResNet34}        &
                 \multicolumn{2}{c|}{$\mathcal{T}:$ResNet34}        &
                 \multicolumn{2}{c}{$\mathcal{T}:$ResNet34}  \\
                & \multicolumn{2}{c|}{$\mathcal{S}:$LeNet-half} & \multicolumn{2}{c|}{$\mathcal{S}:$ResNet18} &
                \multicolumn{2}{c|}{$\mathcal{S}:$ResNet18}&
                \multicolumn{2}{c}{$\mathcal{S}:$ResNet18} \\ \hline
\textbf{Method} & \textbf{FLOPs} & \textbf{Accuracy} & \textbf{FLOPs} & \textbf{Accuracy} & \textbf{FLOPs} & \textbf{Accuracy} & \multicolumn{1}{l|}{\textbf{FLOPs}} & \textbf{Accuracy} \\ \hline
Teacher         & 433K           & 98.9\%            & 1.16G          & 96.3\%            & 1.16G          & 95.4\%            & \multicolumn{1}{l|}{1.16G}           & 94.1\%            \\
Train with data         & 139K           & 98.6\%            & 558M          & 96.0\%            & 558M          & 93.9\%            & \multicolumn{1}{l|}{558M}           & 94.0\%            \\ \hline
RDSKD            & 139K           & 97.6\%            & 558M           & 94.6\%            & 558M           & 90.8\%            & \multicolumn{1}{l|}{558M}           & -                  \\
DAFL            & 139K           & 98.2\%            & 558M           & 94.5\%*            & 558M           & 92.2\%            & \multicolumn{1}{l|}{558M}           & 90.4\%*                  \\
DFAD            & 139K           & \textbf{98.3\%}            & 558M           & 94.7\%*                  & 558M           & \textbf{93.3\% }           & \multicolumn{1}{l|}{558M}           & 70.0\%*                  \\
Ours           & 139K           & 98.2\%            & 558M           & \textbf{95.4\%}            & 558M           & 91.3 \%                 & \multicolumn{1}{l|}{558M}           & 92.3\%                  \\ 
Ours (w/ memory bank)            & 139K           & \textbf{98.3\%}           & 558M           & 95.0\%            & 558M           & 92.4 \%                 & \multicolumn{1}{l|}{558M}           & \textbf{92.9\%}                  \\ \hline
\end{tabular}
\caption{Accuracy results of student networks obtained by different data-free distillation strategies on several datasets. The (*) mark indicates that the results are produced by running the original implementations on previously untested datasets.}
\label{tab:test_results}
\end{table*}


\section{Experiments}
\label{sec:experiments}
In this section, we share the experiments we have conducted to demonstrate the effectiveness of our proposed method on four image classification datasets. Our results are given in Table \ref{tab:test_results} together with those of the baselines, for comparison. Moreover, we discuss how the inclusion of each technique we propose and hyper-parameter choices related to them, affect distillation performance. We consider DAFL \cite{chen2019data}, DFAD \cite{fang2019data}, 
EATSKD \cite{nayak2021effectiveness}, DeGAN \cite{addepalli2020degan}
and RDSKD \cite{han2021robustness} as our baselines for comparison using the same student and teacher models. It is noted, to test the baselines on the benchmarks that they had not reported results on, we used the code on their GitHub repositories.
\subsection{Datasets}
\label{ssec:datasets}
In our experiments, we have used MNIST \cite{lecun1998gradient}, SVHN \cite{netzer2011reading}, CIFAR-10 \cite{krizhevsky2009learning}, 
CIFAR-100 \cite{krizhevsky2009learning}, 
and Fashion MNIST \cite{xiao2017fashion} datasets as benchmarks. In all experiments we use the DCGAN \cite{radford2015unsupervised} inspired generator adopted from \cite{chen2019data}.
\vspace{-10pt}

\begin{figure*}[!ht]
\begin{minipage}[b]{.49\linewidth}
  \centering
  \centerline{\includegraphics[height=.65\textwidth]{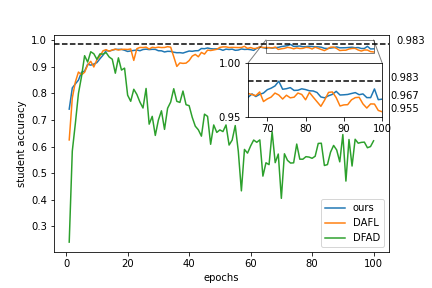}}
  \centerline{(a)}\smallskip
\end{minipage}
\begin{minipage}[b]{.49\linewidth}
  \centering
  \centerline{\includegraphics[height=.65\textwidth]{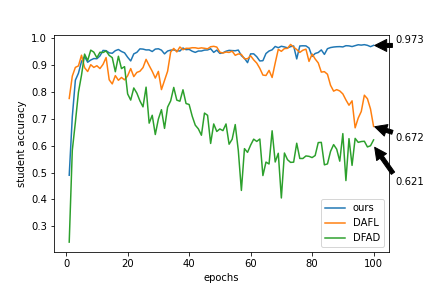}}
   \centerline{(b)}\smallskip
\end{minipage}
\hfill
\caption{Progression of student accuracy throughout distillation achieved by our method in comparison with DAFL and DFAD.}
\label{fig:update_freq}
\end{figure*}

\begin{figure*}[!ht]
\begin{minipage}[b]{.49\linewidth}
  \centering
  \centerline{\includegraphics[height=.65\textwidth]{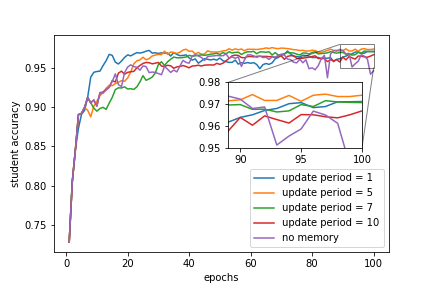}}
  \centerline{(a)}\smallskip
\end{minipage}
\begin{minipage}[b]{.49\linewidth}
  \centering
  \centerline{\includegraphics[height=.65\textwidth]{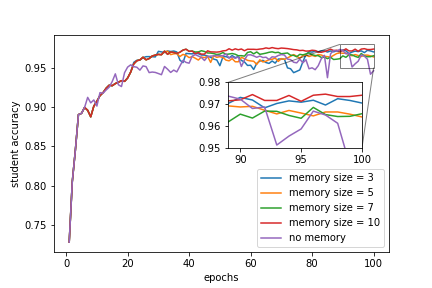}}
   \centerline{(b)}\smallskip
\end{minipage}
\hfill
\caption{Impacts of update rate and memory size on the student training accuracy.}
\label{fig:dataset_size}
\end{figure*}

\paragraph{MNIST} is an image classification dataset composed of 70,000 grey-scale images of handwritten digits from 0 to 9. Training set consists of 60,000 images while the testing set consists of 10,000. This benchmark is simple and therefore student-teacher gaps in most methods are very small. For the experiments on MNIST, we selected LeNet-5 \cite{lecun1998gradient} as the teacher and LeNet-5-Half as the student. The teacher model was achieving 98.9\% accuracy on the test set. Our method performed comparably with the DAFL and DFAD baselines while outperforming RDSKD. Our experiments revealed that when we set the memory size as 10 batches and update frequency as 1 per 5 epochs, the accuracy increases more consistently (see Figures \ref{fig:dataset_size}a and \ref{fig:dataset_size}b). However, we achieved the best results with the update rate of once in every 1 epoch and the memory bank size of 10 batches. Additionally, we plot the learning curves of the student obtained by our method and the baselines, during distillation. The Figures \ref{fig:update_freq}a and \ref{fig:update_freq}b demonstrate 
that
our method 
can reduce the variance among student accuracies achieved throughout epochs. This is desirable in real-life applications since, if there is no evaluation data during the distillation process, it is best if the student model performs at its peak when the process ends. Therefore, maintaining high student accuracy over iterations is as important as achieving it. Based on the plots, our method accomplished these better than the baselines DAFL and DFAD.
\vspace{-10pt}
\paragraph{SVHN} is a colored digit classification dataset consisting of over 600,000 labeled images. This dataset is similar to MNIST with the difference of having RGB channels therefore the benchmark is again relatively simple. For the experiments on SVHN, we selected ResNet-34 \cite{lecun1998gradient} as the teacher and ResNet-18 as the student. The teacher model was achieving 96.3\% accuracy on the test set.  Our method outperformed all baselines. While testing the inclusion of the memory bank, we used an update period of 5 epochs and a memory size of 10 batches. For the remaining benchmarks, we used the same memory bank settings.  
\vspace{-10pt}
\paragraph{CIFAR10} is a colored image classification dataset consisting of over 60,000 labeled images from 10 categories, each category containing 6000 samples. The number of samples ratio for training and testing sets is 5/1. This benchmark contains less salient samples for classification and therefore is more challenging than MNIST and SVHN. For the experiments on CIFAR10, we selected ResNet-34 \cite{lecun1998gradient} as the teacher and ResNet-18 as the student. The ResNet-34 teacher model was achieving 95.4\% accuracy on the test set.  Our method containing the memory bank, outperformed DAFL and RDSKD baselines while performing lower than DFAD. 
\vspace{-10pt}
\paragraph{Fashion MNIST} is a dataset of 70,000 grey-scale cloth images from 10 different categories.  The number of samples in the training and testing sets are 60,000 to 10,000 respectively. To compare our work with DAFL and DFAD, we selected ResNet-34 \cite{lecun1998gradient} as the teacher and ResNet-18 as the student. The ResNet-34 teacher model was achieving 94.1\% accuracy on the test set. 
Our method outperformed both baselines. The improvement our method achieves over Fashion MNIST is much clearer than it achieves over other benchmarks. This could be due to a particular organization of the teacher decision space, tightening the performance bottleneck of the baselines caused by their above-mentioned weaknesses. We note that the results of our baselines we report are obtained after running repeated experiments with different random initializations. Even after numerous trials, DFAD could not maintain the distillation quality it achieved in other benchmarks. We relate this to its frailty against non-optimal hyper-parameter choices, which is also recognized by \cite{han2021robustness}.  

\vspace{-10pt}
\paragraph{CIFAR100} is a more extensive version of CIFAR10 with 100 object categories. There are 600 colored images for each type of object. These total 60,000 samples are split into training and testing sets with a 5 to 1 ratio. For the experiments on CIFAR100, we selected ResNet-34 \cite{lecun1998gradient} as the teacher and ResNet-18 as the student. The ResNet-34 teacher model was achieving 77.94\% accuracy on the test set. Our method containing the memory bank outperformed all baselines. 

\begin{table}[!htb]
\centering
\begin{tabular}{l|l|ll}
\hline
                & \multicolumn{2}{c}{CIFAR100}                       \\ 
                & \multicolumn{2}{c}{$\mathcal{T}:$ResNet34}  \\
                & \multicolumn{2}{c}{$\mathcal{S}:$ResNet18} \\ \hline
\textbf{Method} & \multicolumn{1}{l|}{\textbf{FLOPs}} & \textbf{Accuracy} \\ \hline
Teacher         & \multicolumn{1}{l|}{1.16G}            & 77.94\%            \\
Train with data  & \multicolumn{1}{l|}{558M}          & 76.53\%            \\ \hline
DeGAN            & \multicolumn{1}{l|}{558M}           & 65.25\%                  \\
EATSKD            & \multicolumn{1}{l|}{558M}           & 67.18\%                \\
DFAD            & \multicolumn{1}{l|}{558M}           & 67.70\%                \\
DAFL            & \multicolumn{1}{l|}{558M}           & 74.47\%                \\
Ours            & \multicolumn{1}{l|}{558M}           & \textbf{75.35\%}                  \\ \hline
\end{tabular}
\caption{Accuracy results of student networks obtained by different data-free distillation strategies on CIFAR100 dataset.}
\label{tab:test_results4}
\end{table}

\vspace{-10pt}
\par
In summary, Tables \ref{tab:test_results} and \ref{tab:test_results4} show that our method improved the highest accuracy reached by distilled student networks for SVHN, Fashion MNIST and CIFAR100 datasets. Moreover, although it performs lower than DFAD on CIFAR10, our method consistently maintained high performance across different benchmarks. Considering these, we can claim that our efforts in mitigating catastrophic forgetting and avoiding mismatch between synthetic and real data, can improve the overall quality of data-free KD. 
\subsection{Ablation Study}
\label{subsec:ablation}
To demonstrate how each component in our proposed method impacts the distillation performance, we reported the performances of our method after the inclusion of each component (see Table \ref{tab:test_results}). The addition of the constraint to generate samples that match the original data distribution improved accuracy compared to prior methods for SVHN and Fashion MNIST benchmarks. Later, the inclusion of the memory bank, boosted our performance in all benchmarks except SVHN, yielding comparable results with DFAD on MNIST and better performance than DAFL on CIFAR10. 
\\
\indent Additionally, in Table \ref{tab:ablation}, we compare the results obtained by different choices of divergence terms that force the generator to produce novel samples for the student. For this experiment, we used our method without the memory bank. Choosing JS divergence over other alternatives caused a 
\begin{table}[!htb]
\centering
\begin{tabular}{|l|l|l|l|}
\hline
                 & \multicolumn{1}{c|}{JS} & \multicolumn{1}{c|}{L1} & \multicolumn{1}{c|}{KLD} \\ \hline
Student Accuracy & 92.3\%                  & 92.1\%                  & 92.1\%                   \\ \hline
\end{tabular}
\caption{Effectiveness of different distance terms used to produce novel samples for the student.}
\label{tab:ablation}
\vspace{-10pt}
\end{table}
\\ slight improvement in the performance. We attribute this to the symmetric property of JS divergence as opposed to the asymmetric KL divergence. While the KL divergence quantifies the distance of two distributions from the reference point of one, JS divergence considers both reference points. This is due to the definition of JS divergence given in Equation \ref{eq5_2} which contains a combination of both the KL divergence of the student from the teacher and that of the teacher from the student.
Such way to represent the teacher-student discrepancy could be more suitable than the asymmetric KL divergence.

\subsection{Memory Bank Update Frequency}
\label{sec:bank_update_freq}
The update frequency for the memory bank is a hyper-parameter that needs to be pre-determined and set before distillation. To find the optimal value, we compared the behaviors of different update rates for MNIST dataset distillation. Since the randomness in the parameter initialization could have affected our comparison, we set the random seeds manually. The memory bank size is limited to contain at most 10 batches. We shared the outcome of our experiments in Figure \ref{fig:dataset_size}a. From the plot, it can be observed that regardless of the choice of update rate, the usage of memory bank improved the peak and final values of student accuracy while also causing fewer fluctuations. Moreover, among the update frequencies we tested, 1 update per 5 epochs performed better than others by resulting in a more smooth learning curve and higher student accuracy. Meanwhile, updating the stored samples every epoch, caused larger fluctuations relative to other choices. We attribute this to the rapid refresh rate unable to store samples for enough iterations. On the other hand, the less frequent update periods such as 7 and 10 could be causing failure to memorize some informative samples generated in between updates.   
\subsection{Memory Bank Size}
\label{sec:bank_size}
Selection of the maximum number of batches to be stored in the memory bank is another hyper-parameter choice. To observe the isolated impacts of such choice, we fix the update period to be 5 epochs and plot the learning curves obtained by different memory sizes. We again used manual random seed for a fair comparison. Figure \ref{fig:dataset_size}b shows that storing at most 10 batches at any step throughout distillation resulted both in a higher final student accuracy and a smoother increase in accuracy. Additionally, we observed a correlation between memory size and smoothness of the curve, as keeping fewer samples in memory caused larger fluctuations. We note that any choice of memory size yielded a better performing student than not storing samples. 
\subsection{Distillation Among Different Architecture Types}
\label{sec:dist_diff_arch}
To the best of our knowledge, all prior work had reported results on teacher-student pairs that contain the same type of architectural blocks. For the LeNet5 teacher, the student is typically picked as LeNet-half and for the ResNet34 teacher, ResNet18 is selected as the student. In this work, we also practice distillation among networks with different architectural blocks such as ResNet and MobileNetV2. We used the same ResNet34 teacher trained on Fashion MNIST as in Table \ref{tab:test_results} while changing the student to be MobileNetV2. From Table \ref{tab:test_results2}, it can be observed that our method achieved higher performance than DAFL even 
\begin{table}[!htb]
\centering
\begin{tabular}{l|l|l|ll}
\hline
                & \multicolumn{4}{c}{Fashion MNIST}                       \\ 
                & \multicolumn{2}{c|}{$\mathcal{T}:$ResNet34} & \multicolumn{2}{c}{$\mathcal{T}:$ResNet34} \\
                & \multicolumn{2}{c|}{$\mathcal{S}:$MobileNetV2} & \multicolumn{2}{c}{$\mathcal{S}:$ResNet18} \\ \hline
\textbf{Method} & \multicolumn{1}{l|}{\textbf{FLOPs}} & \textbf{Acc.} & \multicolumn{1}{l|}{\textbf{FLOPs}} & \textbf{Acc.} \\ \hline
Teacher         & \multicolumn{1}{l|}{1.16G}            & 94.1\%  & \multicolumn{1}{l|}{1.16G}            & 94.1\%           \\
Train w/ data  & \multicolumn{1}{l|}{16M}          & 92.4\%  & \multicolumn{1}{l|}{558M}          & 94.0\%          \\ \hline
DAFL            & \multicolumn{1}{l|}{16M}           & 85.7\%* & \multicolumn{1}{l|}{558M}           & 90.4\%*                  \\
Ours            & \multicolumn{1}{l|}{16M}           & \textbf{91.3\%} & \multicolumn{1}{l|}{558M}           & \textbf{92.9\%}                  \\ \hline
\end{tabular}
\caption{Accuracy results of MobileNetV2 student networks obtained by different data-free distillation strategies on Fashion MNIST dataset. The (*) mark indicates that the results are produced by running the original implementations on previously untested settings.}
\label{tab:test_results2}
\vspace{-12pt}
\end{table}
\\
when we change the student network architecture. The results we report of our methods were obtained by including both the memory bank and the proposed generator loss in our KD framework. 

\section{Conclusion \& Future Work}
\label{sec:conclusion}
In this paper, we identify two problems that impair data-free knowledge distillation performance and propose methods to solve them. These are catastrophic forgetting and, mismatch between synthetic and real data distributions. 
\\
\indent Addressing the first problem, we proposed a memory system where we keep the history of generated samples over iterations. While this simple approach manages to suppress the negative effects of catastrophic forgetting on student accuracy, it also increases the secondary memory overhead and slows down distillation. Therefore, we believe future efforts could be directed to find more efficient ways to prevent catastrophic forgetting in data-free KD. Some possible improvements could be achieved by selectively storing samples and/or by storing compressed versions of the samples. 
\\
\indent To remedy the second problem, we propose a data-free KD strategy that constrains the generated novel samples to match the distribution of the original data. This way, we ensure that the synthetic samples distinguishing the teacher from the student are not irrelevant to the original training data. Additionally, to produce such distinguishing samples we adopt Jensen-Shannon divergence as a loss term and compare it with other alternatives. 
\\
\indent Experimental results show that our framework not only improves data-free knowledge distillation performance for certain datasets, but also maintains high student accuracy throughout the entire process.

\section*{Acknowledgement}
This research is partially supported by the National Research
Foundation, Singapore under its Competitive Research Programme Award
NRF-CRP23-2019-0003. We thank Cihan Acar for helpful comments on the manuscript.

\pagebreak
{\small
\bibliographystyle{ieee_fullname}
\bibliography{egbib}
}

\end{document}